\documentclass[sigconf,nonacm]{acmart}
\usepackage[braket, qm]{qcircuit}
\usepackage{soul} 
\setcopyright{none}
\renewcommand\footnotetextcopyrightpermission{} 
\AtBeginDocument{%
  }

\acmISBN{978-1-4503-XXXX-X/2018/06}

\begin{document}

\title{Q-Fusion: Diffusing Quantum Circuits}


\author{Collin Beaudoin}
\email{cpb5867@psu.edu}
\affiliation{%
  \institution{Penn State University}
  \city{University Park}
  \state{Pennsylvania}
  \country{USA}
  \postcode{16801}
}

\author{Swaroop Ghosh}
\email{szg212@psu.edu}
\affiliation{%
  \institution{Penn State University}
  \city{University Park}
  \state{Pennsylvania}
  \country{USA}
  \postcode{16801}
}


\begin{abstract}

Quantum computing holds great potential for solving socially relevant and computationally complex problems. Furthermore, quantum machine learning (QML) promises to rapidly improve our current machine learning capabilities. However, current noisy intermediate-scale quantum (NISQ) devices are constrained by limitations in the number of qubits and gate counts, which hinder their full capabilities. Furthermore, the design of quantum algorithms remains a laborious task, requiring significant domain expertise and time. Quantum Architecture Search (QAS) aims to streamline this process by automatically generating novel quantum circuits, reducing the need for manual intervention. In this paper, we propose a diffusion-based algorithm leveraging the LayerDAG framework \cite{li2024layerdag} to generate new quantum circuits. This method contrasts with other approaches that utilize large language models (LLMs), reinforcement learning (RL), variational autoencoders (VAE), and similar techniques. Our results demonstrate that the proposed model consistently generates 100\% valid quantum circuit outputs.
\end{abstract}





\maketitle

\section{Introduction}

Quantum computing is an emerging field with the potential to significantly advance computational capabilities. At the heart of quantum computing lies quantum circuits, which are analogous to their classical counterparts but operate on qubits to perform complex computations. Quantum gates are the fundamental building blocks of these circuits, with common single-qubit gates including Hadamard (H), Pauli X/Y/Z gates, and rotation gates such as RX, RY, RZ, and U. Multi-qubit gates, including CNOT, Toffoli, controlled Pauli, and controlled rotation gates, enable more intricate operations. Each gate is unitary, which means that it can be represented by a unitary matrix $U$ (where $U^{*}$, $UU^{*}=UU^{-1}=I$), acting on one or more qubits. These operations are crucial for the development of quantum algorithms and the realization of the potential of quantum computing.

Quantum circuits typically consist of a sequence of gate operations applied to various qubits, followed by a measurement of the relevant qubits at the conclusion of the computation. In addition to these functional gate operations, some quantum circuits include a state preparation step prior to the main operations, where the initial state of the qubits is instantiated. This state preparation is crucial for setting up the qubits in the desired configuration before the circuit performs the intended computation.

In the current era of noisy intermediate-scale quantum (NISQ) \cite{preskill2018quantum} devices, quantum computers face several constraints that affect the fidelity and accuracy of quantum circuit executions. These challenges include qubit crosstalk, limited gate types, gate errors, short decoherence times, and restricted qubit connectivity. In these noisy environments, factors such as gate count and circuit depth become critical limitations when designing and evaluating quantum circuits, hindering researchers' ability to develop practical and scalable quantum algorithms. These constraints present a significant barrier to fully realizing the potential of quantum computing.


Even if noise issues were negligible, the high cost of entry and the steep learning curve associated with quantum computing still limit the number of algorithms researchers can develop to harness its unique properties \cite{rubinstein2001evolving, gepp2009review}. Quantum architecture search (QAS) solutions aim to address these challenges by reducing the reliance on human expertise in quantum algorithm development \cite{martyniuk2024quantum}. These approaches employ a variety of techniques, including reinforcement learning (RL) \cite{kuo2021quantum, chen2023asynchronous}, large language models (LLMs)  \cite{liang2023unleashing, dupuis2024qiskit}, variational autoencoders (VAEs) \cite{he2024gradient}, and evolutionary algorithms \cite{rubinstein2001evolving, spector1999finding}. They can be applied to generate both parametric \cite{he2023gnn, du2022quantum} and non-parametric \cite{williams1998automated, yabuki2000genetic} quantum circuits. Parametric Quantum Circuits (PQC), which use quantum gates requiring parametric input, are particularly prominent in Variational Quantum Algorithms (VQAs). VQAs enable the implementation of quantum machine learning models, potentially overcoming the limitations of classical machine learning approaches. In this work, we leverage a diffusion model to generate quantum circuits, aiming to further expand the possibilities of AI designed quantum algorithms for quantum computing.

The structure of the paper is as follows: Section \ref{background} provides the background and related work. Section ~\ref{approach} outlines the approach for converting the models into valid quantum circuit graph generators. Section ~\ref{results} presents visualizations showcasing each model's ability to process quantum circuits. Section ~\ref{discussion} and ~\ref{limitations} offers an analysis of the results, and Section ~\ref{conclusion} concludes the paper.

\section{Background and Related Work} \label{background}
\subsection{Fundamentals of Quantum}
\subsubsection{Qubits} 
Qubits are the fundamental units of a quantum computer, analogous to classical bits in traditional computing. A qubit is typically represented by the quantum state $\ket{\psi}=$
$\big[\begin{smallmatrix}
\alpha \\
\beta
\end{smallmatrix}\big]$, where \( \alpha \) and \( \beta \) are complex numbers. The squared magnitudes of \( \alpha \) and \( \beta \), \( |\alpha|^2 \) and \( |\beta|^2 \), represent the probabilities of the qubit being measured as classical 0 or 1, respectively. These probabilities must sum to 1, i.e., \( |\alpha|^2 + |\beta|^2 = 1 \). The two primary states of a qubit are $\ket{0}$ (where \( \alpha = 1 \) and \( \beta = 0 \)) and $\ket{1}$ (where \( \alpha = 0 \) and \( \beta = 1 \)). Qubit states can be manipulated using unitary gates, and upon measurement, the qubit collapses into one of these two classical states, 0 or 1.

\subsubsection{Quantum Gates} 
Quantum gates are unitary matrix operations that manipulate the quantum states of one or more qubits. The implementation of quantum gates varies depending on the qubit technology used, with different physical systems realizing gates through distinct methods. For example, Nuclear Magnetic Resonance (NMR) qubits utilize radio frequency pulses \cite{gershenfeld1998quantum}, while superconducting qubits rely on microwave pulses \cite{krantz2019quantum}. In contrast, trapped ions and quantum dots use laser pulses to perform gate operations \cite{benhelm2008towards,nguyen2012optically}. The operation speed of these gates can vary significantly between different technologies, with NMR qubit gates that take several seconds to execute, while photonic qubit gates can operate on the order of picoseconds \cite{ladd2010quantum}.

A few examples of single-qubit gates are the: rotational gates like $R_X$/$R_Y$/$R_Z$, X (NOT) gate, $U$ gate, and the H (Hadamard) gate. Gates that control multiple qubits consist of the Toffoli gate, the Peres gate \cite{peres1985reversible}, the CNOT gate (controlled-NOT) and controlled rotation gates.

\subsection{Diffusion Neural Networks}

\begin{figure}[t]
     \centering
         \includegraphics[width=0.75\linewidth]{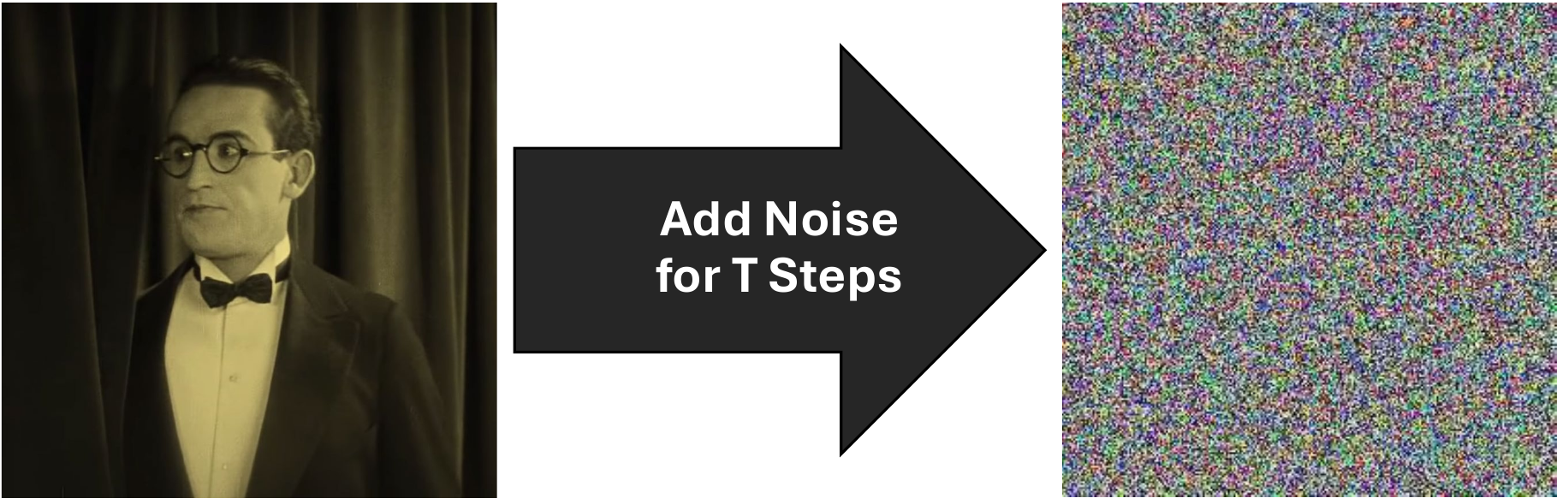}
         \includegraphics[width=0.75\linewidth] {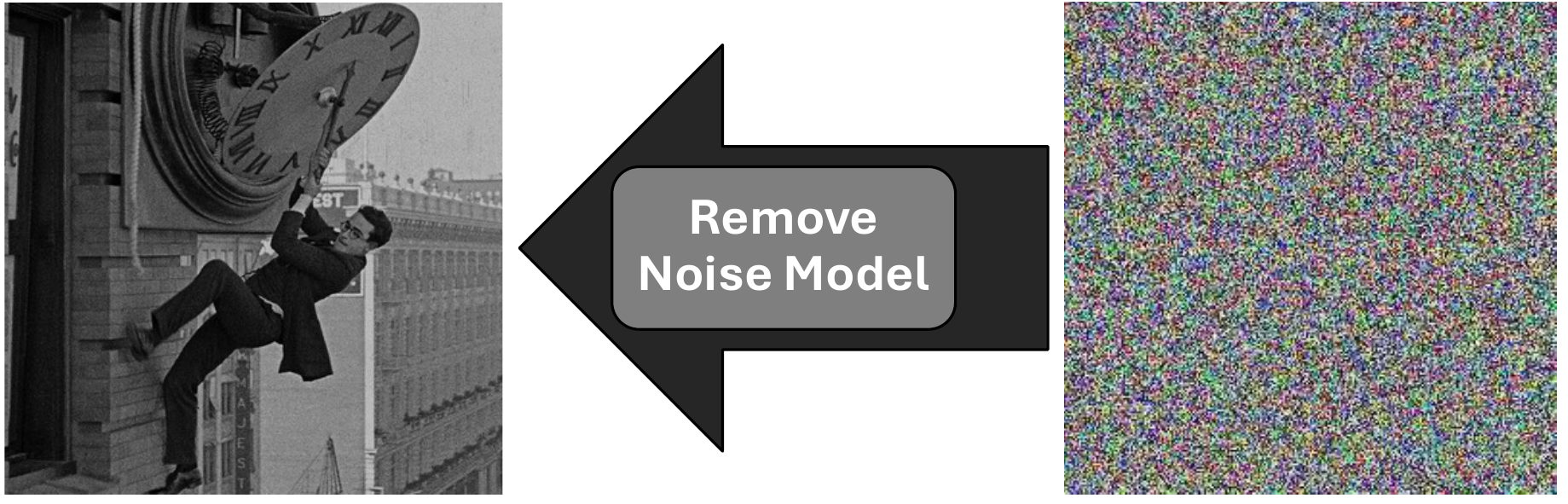} 
        \caption{Sample of the diffusion process. The top shows how an image structure is destroyed via noise over several steps. The bottom shows how structure is recreated using a model.
        }
        \label{fig:sample_diffusion}
\end{figure}

Diffusion models are widely used in image generation to create or recover data samples that align with the distribution of source data \cite{sohl2015deep}. The diffusion process consists of two phases: a forward diffusion phase and a reverse diffusion phase. In the forward phase, noise is incrementally added over several time steps, progressively obscuring the underlying structure of the data. Once the data is sufficiently corrupted by noise, it is passed through a model during the reverse diffusion phase. Here, the model is trained to learn the noise patterns so it can reverse the process, gradually denoising the data to reveal the original structure. Diffusion models have gained significant popularity in recent years, surpassing Generative Adversarial Networks (GANs) in certain applications. They have also been successfully integrated with multi-modal models to generate images based on textual descriptions \cite{dhariwal2021diffusion, rombach2022text, croitoru2023diffusion} (Figure ~\ref{fig:sample_diffusion}).

\subsection{Quantum Architecture Search} 
Current quantum architecture search (QAS) solutions face several challenges. Large language model (LLM)-based approaches, which leverage vast datasets and are central to neural network research, such as the Qiskit code assistant, have the potential to enable researchers to design quantum circuits more efficiently \cite{dupuis2024qiskit}. However, LLMs lack the ability to fully exploit the physical structures defined by graph-based approaches, which can lead to inaccurate or suboptimal results due to the absence of physical understanding \cite{liang2023unleashing}. Reinforcement learning (RL)-based solutions allow for a more thorough exploration of the quantum search space using various techniques \cite{kuo2021quantum, mckiernan2019automated, pirhooshyaran2021quantum}, but they typically rely on expert-defined sub-circuit replacement rules, making them time-consuming to train and run. Variational autoencoder (VAE)-based solutions, while able to leverage the graph structure of quantum circuits and avoiding the training/runtime inefficiencies of RL, still face scalability issues and struggles with qubit size \cite{he2024gradient}. Moreover, VAEs are prone to problems such as mode collapse and unstable training. To overcome these limitations, we propose a diffusion model based on the LayerDAG framework \cite{li2024layerdag} as a novel approach for quantum circuit generation.

\section{Q-Fusion: Approach} \label{approach}
\subsection{Approach}

To harness the generative potential of diffusion models in the quantum domain, we introduce Q-Fusion, an innovative approach to generate quantum circuits. Our method builds upon quantum circuits' natural representation using graph notation and draws inspiration from recent graph-based diffusion frameworks, such as LayerDAG \cite{li2024layerdag}. In this graph representation, nodes correspond to quantum gates, whereas edges represent the connections (wires) between qubits. In this section, we detail the functionality of LayerDAG and the necessary modifications to adapt it for the effective creation of Q-Fusion, enabling the generation of valid and efficient quantum circuits.

\begin{figure*}[t]
    \centering
    \includegraphics[width=.75\linewidth]{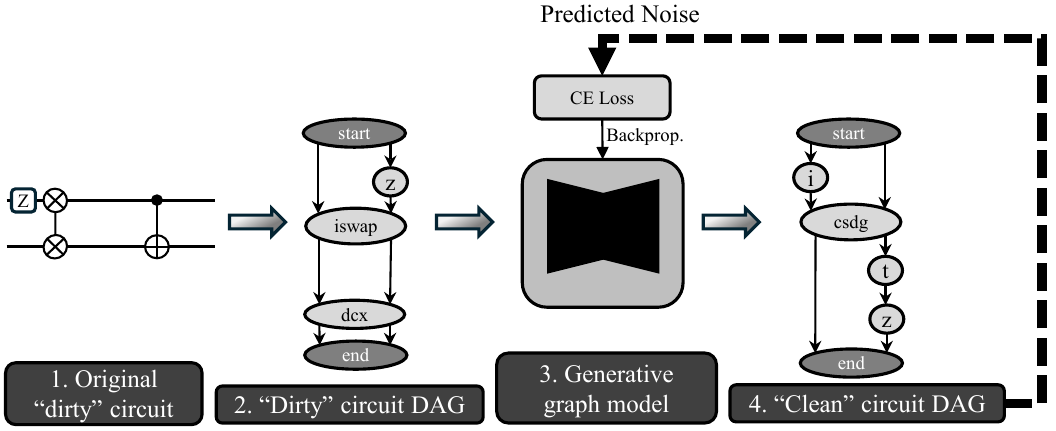}
    \includegraphics[width=.75\linewidth]{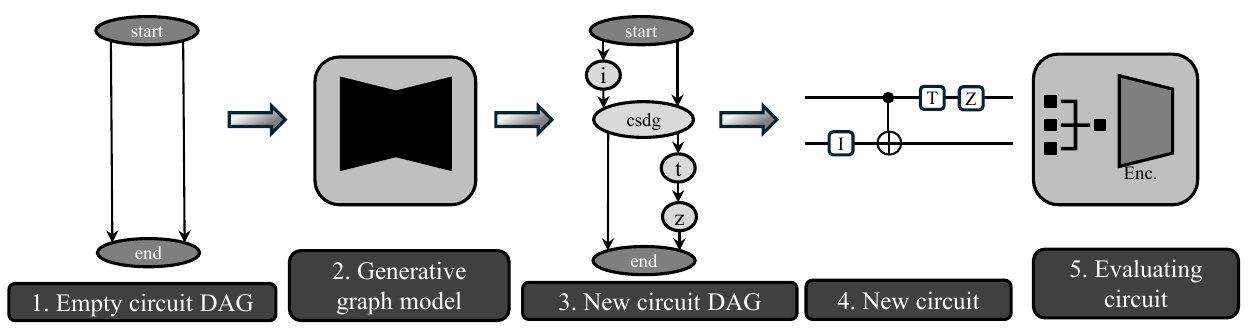}
    \caption{Overview of Q-Fusion process for generating quantum circuits. The top image represents the training process of the diffusion model. The bottom image displays the sampling method used to create new circuits from the trained diffusion model. }
    \label{fig:qfusion}
\end{figure*}

The process of generating quantum circuits using Q-Fusion is illustrated in Fig. \ref{fig:qfusion}. Initially, the original quantum circuits (step \raisebox{0.5pt}{ \textcircled{\raisebox{-0.9pt}{1}}}) are transformed into directed acyclic graphs (DAGs), with noise added to "dirty" the graph (step \raisebox{0.5pt}{ \textcircled{\raisebox{-0.9pt}{2}}}). These "dirty" DAGs are then input into the graph diffusion model (step \raisebox{0.5pt}{ \textcircled{\raisebox{-0.9pt}{3}}}), which reconstructs "clean" DAGs (step \raisebox{0.5pt}{ \textcircled{\raisebox{-0.9pt}{4}}}). The noise from this process is used to update the model. During training, we employ Cross-Entropy (CE) Loss to compare the real noise applied with the predicted noise, ensuring that the model learns to preserve the desired functionality. In the sampling phase, empty DAGs are generated (step \raisebox{0.5pt}{ \textcircled{\raisebox{-0.9pt}{1}}}) and fed into the diffusion model (step \raisebox{0.5pt}{ \textcircled{\raisebox{-0.9pt}{2}}}). The model then samples to produce new, completed graphs (step \raisebox{0.5pt}{ \textcircled{\raisebox{-0.9pt}{3}}}). Finally, the generated graph is converted into a quantum circuit (step \raisebox{0.5pt}{ \textcircled{\raisebox{-0.9pt}{4}}}), which is validated for correctness as a quantum circuit (step \raisebox{0.5pt}{ \textcircled{\raisebox{-0.9pt}{5}}}).

\subsection{Training} \label{training}

\begin{figure}[t]
     \centering
         \includegraphics[width=0.5\linewidth]{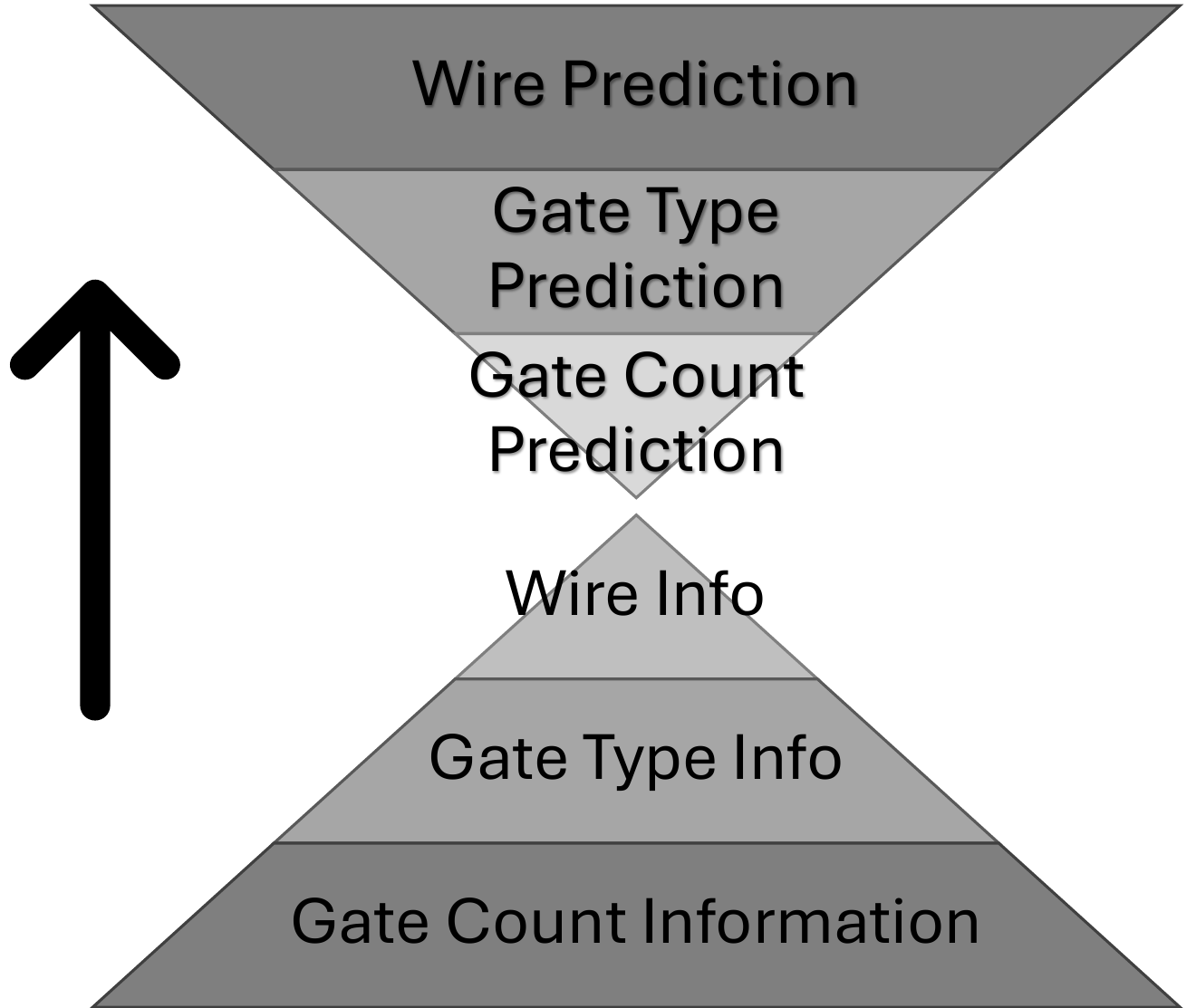}
        \caption{Underlying structure of 3 layered diffusion model. 
        }
        \label{fig:qfusion-structure}
    \vspace{-4mm}
\end{figure}

LayerDAG employs a three-pronged layering approach, visualized in Figure ~\ref{fig:qfusion-structure}, with each layer trained using independent parameters. The first layer predicts the node count, the second layer predicts the nodes themselves, and the third layer predicts the edges between the nodes. Each layer utilizes a variation of teacher forcing, where the expected label is fed into the graph encoding portion of the model during training. In the context of quantum computing, several labeling strategies are possible. After experimenting with various options, we chose to use the summation of the density matrix as the label. This choice stems from the sparse nature of most density matrices, where we sacrifice some expressivity to achieve a significant reduction in memory usage (from $O(2^n * 2^n)$ to $0(1)$).

For each of the three layers, the directed acyclic graph (DAG) is processed sequentially. A noisy partial DAG is fed into the model, which then predicts the information for the next layer (i.e., the node count, the nodes themselves, and the edges for that layer). The parameters of each model layer are updated using Cross-Entropy (CE) Loss, comparing the predicted noise with the actual noise applied to the model, as is common in modern diffusion models \cite{austin2021structured}. This approach ensures that each layer progressively refines the structure of the DAG toward a valid quantum circuit.

\subsection{Sampling}

Once the model, with all its separate layers, has been trained, it is sampled for performance evaluation. During the sampling process, empty graphs with noise are fed into the model, which predicts the features of the next layer—specifically, the number of nodes, node types, and edges. The predictions from each layer are then used to update the graph, and this process is iterated until the model predicts no new nodes. Upon completion, we attempt to convert the resulting DAGs into quantum circuits. The model’s performance is then evaluated based on its ability to generate graphs that align with valid quantum circuits, ensuring that the generated structures are both functional and meaningful in the context of quantum computation.

\subsection{Model Updates}

\subsubsection{Graph Generation} 

LayerDAG is designed to process graphs without requiring a standardized structure, which is advantageous in its ability to handle generic directed acyclic graphs (DAGs). However, quantum circuits have a highly structured flow of information, governed by strict layering and connection rules. For example, altering the order of gate execution or removing a gate can drastically change the functionality of a circuit. Similarly, if a graph were to bifurcate, the circuit could fail, the qubit count might be reduced, or the execution order could become unpredictable. Dropping a qubit from the structure could render the circuit non-functional or fail to preserve its original design. To address these challenges, we introduce virtual start and end nodes at the beginning and end of the training graphs. These nodes help guide the diffusion model by emphasizing the importance of maintaining the strict structure and order of execution inherent in quantum circuits. This approach aims to ensure that the generated circuits remain consistent with the intended functionality and quantum operations.

Ideally the quantum circuits used would come from a large, well maintained dataset where the circuits are labeled for their functionality. Unfortunately, in the NISQ era, this is not available to the public. So to build a data set, random quantum circuits are generated, and then converted to graphs to train the diffusion model.

\subsubsection{Wire Information}

While LayerDAG is capable of processing generic DAGs, it may overlook critical structural details. In the case of quantum circuits, for instance, there is no inherent way to represent which wires are associated with the edges, making it difficult to determine how the circuit functions when multiple qubits are involved. As the number of qubits increases, the number of potential circuit configurations grows factorially, further complicating the task. The absence of wire information also removes key structural elements that help define the quantum search space. To address this, we introduce wire information into each of the three layers of the model. This is accomplished by adding additional parameters that build a hidden representation of the wire connections, similar to the way node types are represented. These wire parameters are then concatenated with the overall hidden representation of the graph before being processed to make predictions. This addition ensures that the model can better capture the intricate connectivity and structure of quantum circuits, leading to more accurate and functional predictions.

\subsection{Model Information}
\subsubsection{Gates \& Gate Connections}

The model employs topological ordering for both graph encoding and reconstruction. During the reconstruction phase, it allows for the selection of edges between the current nodes and any previously created nodes. The model is also trained to learn the appropriate number of incoming and outgoing edges based on the types of gates involved. While quantum gates have fixed input and output edge constraints, we designed the model to explore this space and autonomously learn the structural limitations. This approach enables the model to discover the optimal configurations for gate connections. When the model predicts that no new nodes remain for a given layer, it halts the graph processing, signifying the completion of the quantum circuit structure.

\subsubsection{Circuit Types} \label{circ_types}

As a proof of concept to evaluate Q-Fusion's ability to produce quantum circuits a large set of non-parametric gates using the $X, Y, Z, H, S, T, ID, SXDG, SDG, SX,$ and $TDG$ single qubit gates and the $CX$, $CY$, $CZ$, $SWAP$, $DCX$, $ISWAP$, $CSDG$, $ECR$, $CH$, $CS,$ and $CSX$ were used to generate 6k samples of 2 qubit 8 gate circuits.

To reduce the complexity of the quantum search space and better emulate the constraints of NISQ machines, the gates used in the design of training and testing graphs are drawn from non-parametric circuits consisting of one- and two-qubit gates from the IBM Heron gate set. Specifically, we utilize single-qubit gates such as \( X \), \( SX \), \( ID \), and the two-qubit gate \( CZ \). Additionally, we evaluate the performance of parametric circuits by incorporating the parametric gate \( RZ \), which allows for more flexible gate operations. This combination of non-parametric and parametric gates facilitates a more controlled exploration of the quantum circuit design space, reflecting the practical limitations of current quantum hardware.

\section{Results} \label{results}

\subsection{Setup}

The Q-Fusion model has been enhanced over LayerDAG \cite{li2024layerdag} with additional structures, including the ability to predict the qubit wires assigned to gates and the connection constraints between gates, to address the specific requirements of quantum circuits. For an initial evaluation, Q-Fusion is tested using a 2-qubit, 8-gate circuit, with 6,000 samples used for model training. The circuit generation utilized the custom non-parametric gate set described in Section ~\ref{circ_types}. During evaluation, the model's wire predictions are excluded, and the virtual start and end nodes are used to guide the graph generation. Wire ordering is randomly assigned at the initial connection from the virtual start node and maintained throughout the graph-to-circuit translation process.

\subsection{Generation of Non-Parametric Circuits}

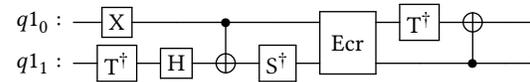
\begin{figure}[h]
\centering
\Qcircuit @C=1.0em @R=0.2em @!R {
  \nghost{{q1}_{0} :} & \lstick{{q1}_{0} :} & \gate{\mathrm{X}} & \qw & \ctrl{1} & \qw & \multigate{1}{\mathrm{Ecr}} & \gate{\mathrm{T^\dagger}} & \targ & \qw & \qw \\
  \nghost{{q1}_{1} :} & \lstick{{q1}_{1} :} & \gate{\mathrm{T^\dagger}} & \gate{\mathrm{H}} & \targ & \gate{\mathrm{S^\dagger}} & \ghost{\mathrm{Ecr}} & \qw & \ctrl{-1} & \qw & \qw
}
\caption{Example Quantum Circuit produced by Q-Fusion for the 2 qubit 8 gate trained model}
\label{fig:ex-circ}
\end{figure}

An example of a generated 2-qubit quantum circuit is shown in Figure ~\ref{fig:ex-circ}. This demonstrates that the diffusion model is capable of effectively navigating the quantum search space, accurately selecting the correct number of wires per gate and distributing gates across multiple wires to construct a functional and unique circuit.

The Q-Fusion model's ability to predict the wire assignment for each gate was reintroduced and evaluated using two smaller circuit sizes: 2 qubits and 5 qubits, each consisting of 32 non-parametric gates. This reduced circuit architecture facilitates the verification of circuit designs without requiring extensive computational resources or runtime for evaluating the density matrices. To train the model, 6,000 random sample circuits were generated for each separate circuit set. The 32-gate circuits were constructed using IBM Heron's non-parametric gate set.

\begin{table}[t]
\caption{Non-Parametric Circuits Produced (\%) From Sampling Trained Diffusion Model}
\label{tab:stats}
\resizebox{\linewidth}{!}{
\begin{tabular}{@{}c|c|cc@{}}
\toprule
\textbf{Statistic} & \textbf{\% Valid} & \textbf{\% Unique} \\
\hline
2-qubit Circuits, Custom & 100\% & 40.21\% \\
2-qubit Circuits, Custom w/ Meaningful Functionality & 99.9\% & 40.21\% \\
2-qubit Circuits, Heron & 100\% & 37.25\% \\
2-qubit Circuits, Heron w/ Meaningful Functionality & 48.92\% & 76.51\% \\ 
5-qubit Circuits, Heron & 100\% & 9.77\% \\
5-qubit Circuits, Heron w/ Meaningful Functionality & 21.89\%  &32.71\% \\ 
\bottomrule
\end{tabular}
}
\end{table}

After successfully training Q-Fusion, the model is sampled to generate 4,320 random circuits in order to assess the ability of the diffusion model to reproduce quantum circuits. The results are presented in Table \ref{tab:stats}. The model demonstrated the ability to generate a circuit successfully each time it is sampled. To evaluate the model's effectiveness in producing "meaningful" circuits (i.e., circuits that are not empty), we analyzed the density matrices of each generated circuit. These density matrices vary in size, with 4x4 matrices containing 16 values and 25x25 matrices containing 625 values. A circuit is considered "meaningful" if 10 or more of these values were non-zero, indicating that the circuit exhibited functional quantum behavior.


\subsection{Generation of Parametric Circuits}

After demonstrating that Q-Fusion can effectively work with quantum circuits, we next evaluated its ability to generate parametric quantum circuits (PQCs) that are fundamental to quantum neural networks (QNNs), the results of which are in Table ~\ref{tab:prob-stats}. PQCs enable the training and tuning of quantum circuits for machine learning applications. For this evaluation, we generated 6,000 random sample circuits, each using between 1 and 5 qubits and comprising 32 gates. The parametric gates were assigned random values between 0 and \(2\pi\). The specific values used for the parameters are not critical, as their precise selection holds limited value without knowledge of the downstream task. These parameters are fine-tuned during the neural network training process, which is where their true utility is realized. 

\begin{table}[t]
\caption{PQCs Produced (\%) From Sampling Trained Diffusion Model}
\label{tab:prob-stats}
\resizebox{\linewidth}{!}{
\begin{tabular}{@{}c|c|c|cc@{}}
\toprule
\textbf{Statistic} & \textbf{\% Valid} & \textbf{Expressibility} & \textbf{\% Unique} \\
\hline
PQCs Produced From Sample Space & 100\% & .953 & 16.94\% \\
PQCs w/ Meaningful Functionality & 32.45\% & .880 & 59.12\% \\ 
\bottomrule
\end{tabular}
}
\end{table}

\section{Discussion} \label{discussion}

\subsection{Non-Parametric Circuit Design}

When evaluating the diffusion model's ability to generate quantum circuits, we observe promising results. However, a closer examination of Table ~\ref{tab:stats} reveals a concerning trend: a decrease in "meaningful" circuits. This decline can be attributed to several factors, with the most significant being the limited gate count and gate types. The circuit is constrained to producing only 32 gates per instance using IBM Heron. As discussed in Section ~\ref{circ_types}, the inclusion of the \(ID\) gate, which functions as a null operation, exacerbates this issue. Moreover, with such a restricted gate set, there is a high likelihood of repeating gate operations on the same wire in rapid succession. This is problematic, particularly for gates like the \(X\) gate, where consecutive applications result in cancellation, effectively behaving like a null operation.

Another contributing factor is the diffusion model's lack of inherent quantum knowledge. While the model understands how to construct a valid layout, select appropriate wiring, and so on, it lacks the built-in understanding of optimization principles like gate cancellation. This absence of quantum awareness further explains the reduction in "meaningful" functionality observed in the results. Addressing this gap in quantum knowledge presents an intriguing avenue for future research, with the potential to improve the model's performance in generating functional quantum circuits.

Additionally, we evaluate the model's ability to produce unique circuits. For the case of the non-meaningful circuits, the 2-qubit custom gate set achieves the highest percent uniqueness. The 2-qubit Heron set obtains the highest value in terms of achieving meaningful functionality. This is due to the fact that the model has a much larger set of gates available to the system allowing for better performance in the case of the 2-qubit custom gate set. However, when we account for the meaningful nature behind the circuits the 2-qubit Heron set performs the best, this is because the smaller gate set and circuit size allows for the model to build a stronger understanding of the nature of quantum circuit design. 

\vspace{-4mm}
\subsection{Parametric Circuit Design} \label{pqc}

Table ~\ref{tab:prob-stats} tabulates an additional statistic, expressibility \cite{sim2019expressibility}, which defines how well a PQC may be able to learn. Values close to 0 are preferred, with no upper limitation on the value observed. The diffusion model is able to produce near 0 values, and we note that the expressibility improves with more meaningful circuits. The performance of LayerDAG is promising, especially considering the expressibility is not considered at training time. Further improvements would be expected if this is taken into account.

Comparing the uniqueness performance between Table ~\ref{tab:stats} to Table ~\ref{tab:prob-stats}, we note that the PQC model's performance sits right between the 2-qubit and 5-qubit non-parametric model. This is expected as the gate set is only increased by a single gate, while consisting of circuits using 1 to 5 qubits.

Continuing to examine Table ~\ref{tab:prob-stats}, we observe a continued loss in "meaningful" functionality. While the issues identified earlier persist, two additional concerns arise. The first relates to the behavior of the parameters. Since parameter values are randomly generated, they may either cancel out the functionality on the wire or leave the wire with no functionality at all. If the \(RZ\) gate is assigned a value that results in the inverse of the wire's intended functionality, the operation effectively becomes a null operation. Given that the only current methods for evaluating a PQC rely on training and assessing its performance on a machine learning task (IE tuning the parameter), this loss of "meaningful" functionality should be regarded as a limitation of quick evaluation techniques. Future work should focus on improving methods for evaluating the functionality of PQCs more reliably.

The second issue contributing to the performance drop stems from the use of a functionally smaller dataset. This choice was deliberate, aiming to better simulate the constraints of quantum systems in the current NISQ era. While previous experiments utilized 6k samples for a single circuit size, in this study, the circuit range was expanded to include 1, 2, 3, 4, and 5 qubit circuits, along with an additional complex parametric gate, all while maintaining the same dataset size. The challenge is that there are not 6k well-understood circuits available, even for these small circuit sizes. Nonetheless, it is important to highlight that meaningful machine learning progress can still be made within this limited data regime, showcasing the potential for developing effective quantum models despite these constraints.

\section{Limitations}

\subsection{Evaluation} \label{limitations}

Diffusion models have demonstrated strong performance with large and complex images, offering promising prospects for scaling quantum circuits to incorporate more gates and qubits. However, these models are not free of limitations, and standard challenges associated with diffusion models and graph representations must be considered. A unique constraint in the context of quantum computing is the difficulty in calculating the density matrix associated with the circuit. Density matrices, which represent the functionality of quantum circuits, scale exponentially as \(2^{n} \times 2^{n}\), where \(n\) is the number of qubits. As circuit sizes increase, this exponential growth quickly outpaces the computational power of classical machines, making it infeasible to represent and evaluate large quantum circuits. For Q-Fusion, this means that, beyond a certain scale, it will no longer be possible to evaluate the circuit design using classical methods. One potential solution is to map the circuit to a physical quantum device, though this would limit our ability to assess its "meaningful" functionality using the current definition. As discussed in Section ~\ref{pqc}, evaluating PQCs in this manner may be unreliable. Nevertheless, if preserving the evaluation of meaningful functionality is critical for the task at hand, a feasible approach could involve decomposing larger circuits into smaller sub-circuits and evaluating them individually.

\subsection{Labeling}

One limitation of the current approach arises from the use of the label system discussed in Section ~\ref{training}). If this label were to be replaced with a density matrix representation, the model would struggle to process large circuit designs. Even with the existing label system, the model becomes unfeasible when the computer can no longer process the entire density matrix. This limitation stems from the fact that the label relies on a summation of the density matrix. However, the ultimate goal of the label is to provide a unique representation for functionally distinct circuits. To overcome this constraint and preserve similar functionality, an alternative approach could involve running the circuit on real hardware and using the resulting output vector to represent the circuit.

\section{Conclusion} \label{conclusion}

Based on the results from both non-parametric and parametric quantum circuit generation as benchmarks, we believe that diffusion models hold significant potential to enhance QAS solutions. In the future, the proposed approach may facilitate the accelerated exploitation of quantum properties, offering promising advancements for complex problems such as quantum machine learning.

\begin{acks}
We acknowledge the usage of IBM Quantum and Pennylane for our experiments. 
\end{acks}

\bibliographystyle{ACM-Reference-Format}
\bibliography{sample-base}

\appendix

\end{document}